\documentclass[sigconf]{acmart}

\usepackage{color,soul}
\usepackage[textsize=footnotesize,textwidth=1cm]{todonotes}
\usepackage{gensymb}
\usepackage{calligra}
\usepackage{amsmath}
\usepackage{bbm}

\AtBeginDocument{%
  \providecommand\BibTeX{{%
    \normalfont B\kern-0.5em{\scshape i\kern-0.25em b}\kern-0.8em\TeX}}}

\setcopyright{acmcopyright}
\copyrightyear{2021}
\acmYear{2021}
\acmDOI{10.1145/xxxxxxx.xxxxxxxx}

\acmConference[KDD-MLF '21]{4th KDD Workshop on Machine Learning in Finance (KDD-MLF '21)}{August 14--18, 2021}{Singapore}
\acmBooktitle{4th KDD Workshop on Machine Learning in Finance (KDD-MLF '21),
  August 14--18, 2021, Singapore}

\begin{document}

\title{Heterogeneous Ensemble for ESG Ratings Prediction}



\author{Tim Krappel}
\email{krappeltim@gmail.com}
\affiliation{%
  \institution{University of St. Gallen}
  \city{St. Gallen}
  \country{Switzerland}
}
\author{Alex Bogun}
\email{alex.bogun@unisg.ch}
\affiliation{%
  \institution{University of St. Gallen}
  \city{St. Gallen}
  \country{Switzerland}
}
\author{Damian Borth}
\email{damian.borth@unisg.ch}
\affiliation{%
  \institution{University of St. Gallen}
  \city{St. Gallen}
  \country{Switzerland}
}


\renewcommand{\shortauthors}{Krappel, Bogun, Borth}

\begin{abstract}
    Over the past years, topics ranging from climate change to human rights have seen increasing importance for investment decisions. 
Hence, investors (asset managers and asset owners) who wanted to incorporate these issues started to assess companies based on how they handle such topics. 
For this assessment, investors rely on specialized rating agencies that issue ratings along the environmental, social and governance (ESG) dimensions.
Such ratings allow them to make investment decisions in favor of sustainability.
However, rating agencies base their analysis on subjective assessment of sustainability reports, not provided by every company.
Furthermore, due to human labor involved, rating agencies are currently facing the challenge to scale up the coverage in a timely manner.

In order to alleviate these challenges and contribute to the overall goal of supporting sustainability, we propose a heterogeneous ensemble model to predict ESG ratings using fundamental data. This model is based on feedforward neural network, CatBoost and XGBoost ensemble members.
Given the public availability of fundamental data, the proposed method would allow cost-efficient and scalable creation of initial ESG ratings (also for companies without sustainability reporting).
Using our approach we are able to explain 54\% of the variation in ratings ($R^2$) using fundamental data and outperform prior work in this area.

\end{abstract}





\maketitle
\section{Introduction}
\begin{figure}[h]
    \centering
    \includegraphics[width=\linewidth]{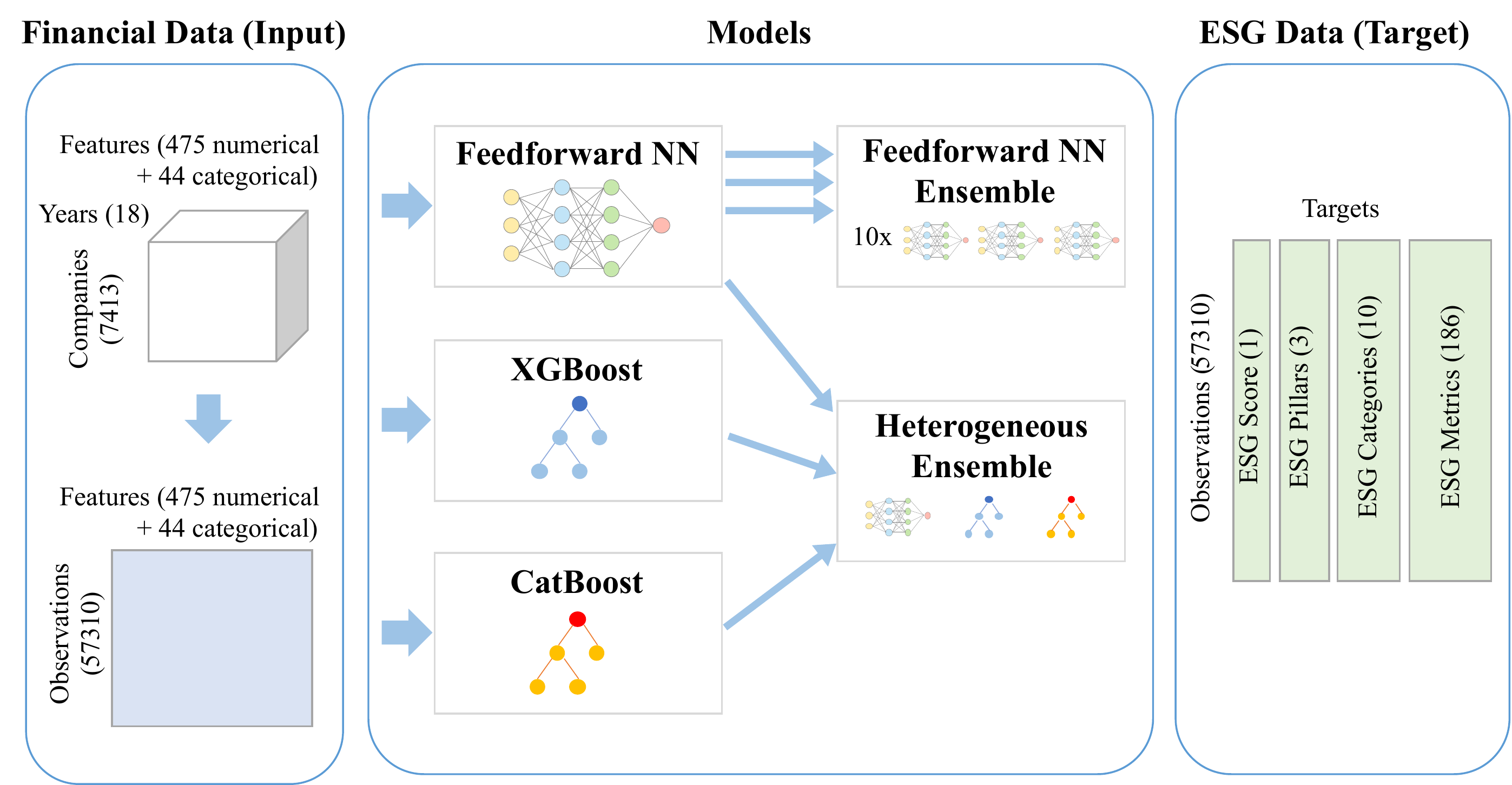}
    \caption{Overview of proposed ESG scores prediction using fundamental financial data. Tree individual models and two types of ensembles (of 10 NNs and of 3 individual models) are used.}
    \label{fig:setup}
    \vspace{-0.5cm}
\end{figure}

If it is not tackled now, climate change will have a large impact on the world \cite{nasa_effects}. With 19 out of the 20 hottest years on record having occurred since 2001 \cite{nasa2020}, countries across the planet have adopted countermeasures. In 2015, parties to the Paris Agreement set out the goal to keep the temperature increase compared to pre-industrial levels below 2\degree C \cite{paris:2015}. Earlier, the sustainable development goals were introduced by the United Nations. These outlined objectives for the international community such as promoting sustainable energy and gender equality or combating climate change \cite{un:2015}. Pension funds, foundations, endowments and other asset owners started incorporating them into strategic asset allocation and mandating their asset managers to track sustainability objectives. In some cases investors were even obligated to do so by new regulations such as the Non-Financial Reporting Directive (NFRD) \cite{eu_nfrd_2014} and the  directive on the Activities and Supervision of Institutions for Occupational Retirement Provision (IORPs) \cite{eu_iorp_2016} in the EU.

These developments have led to the trend of sustainable investing. It marks a shift from an investors’ focus on short-term profits to the consideration of long-term impacts and the non-financial risk of businesses. They are categorized as environmental (E), social (S) and governance (G) issues \cite{anton2020} and relate to topics such as a company's contribution to climate change (E), whether it respects human rights (S), or the rights of shareholders (G) \cite{robeco}. Over the past years, the number of investors who integrate them into their investment decisions has grown. For instance, the number of signatories of the Principles for Responsible Investment (PRI) has increased. These signatories incorporate ESG issues into their decisions and actively promote the disclosure of data related to them \cite{pri}. In 2019, investors that have signed the PRI represented a total of USD 85 trillion assets under management \cite{fid:2019}.

However, in order to incorporate ESG issues into investment decisions, one has to have access to sufficient information about how a company handles them. This may be difficult for some shareholders \cite{chatter2009}. Therefore, organisations emerged which specialised in assessing companies based on how they handle ESG issues and assigning ratings, i.e., ESG ratings to them \cite{cormier2011}. Some of the largest providers include MSCI\footnote{https://www.msci.com/esg-ratings}, Refinitiv\footnote{https://www.refinitiv.com/en/financial-data/company-data/esg-data}, Bloomberg\footnote{https://www.bloomberg.com/professional/solution/sustainable-finance/} or Sustainalytics\footnote{https://www.sustainalytics.com/esg-data/}. This paper uses the data by Refinitiv. Its data and its procedure are explained in section \ref{esgdata}. Since the methodologies of the providers are part of their intellectual property, only an overview can be shown.\par

Over the past years, ESG ratings have seen increasing criticism. Since most countries have not implemented disclosure requirements for companies' handling of ESG issues \cite{fid:2019}, the ratings rely on self-disclosed information. This has led to a number of difficulties. For instance, due to the different composition and weighting of ESG ratings, final ratings often differ substantially for the same companies across providers. Additionally, large corporations often have better ratings since they are able to offer enhanced reporting compared to their smaller competitors \cite{dorf:2015} of which more than 50\% do not report on sustainability at all \cite{martin_survey_2021}. It is difficult to compare ratings from different organisations as the methodology of the rating process is not transparent \cite{scalet2010} and it involves subjective assessment \cite{fowler2007, stubbs2013}. Furthermore, it was shown that ratings seem not to be optimally using public data \cite{chatter2009} and that high ratings in one category might offset low ratings in another \cite{escrig2014}.\par

These problems motivate the creation of a procedure that can infer ESG ratings from regulated and publicly available data. The predicted ESG ratings could be used for smaller companies not covered by rating agencies, companies that do not fill out sustainability or integrated reports, companies shortly after initial public offerings, etc. This could benefit sustainable investors, by allowing them to include companies without ESG rating or with poor coverage into their investment portfolio as well as rating agencies by giving the ability to compute initial ratings quickly. This would contribute to the overall goal of supporting sustainability in the investment industry. This study provides a way for the creation of initial ESG ratings using machine learning models. It is worth noting that ESG controversies (e.g. scandals, litigations, etc.) are not part of our analysis in the same manner as they are not part of ESG scoring by rating agencies.

The contributions of this work can be summarized in the following 3 points:
\begin{itemize}
    \item We introduced a heterogeneous ensemble model based on feedforward neural network, CatBoost and  XGBoost algorithms to predict ESG ratings using only publicly available fundamental data without relying on sustainability reports. To our knowledge, this work is the first to employ advanced machine learning methods for ESG rating prediction.
    \item We outperformed the previous linear regression baseline achieved by \citeauthor{garcia:2020} \cite{garcia:2020}, increasing test accuracy as measured by $R^2$ from 27.4\% to 53.9\% and reducing mean absolute error to 11.2 percentage points.
    \item Our approach allows for the creation of initial ESG ratings. These can be used either directly by investors to evaluate not-rated companies or by ESG rating agencies to quickly scale up their coverage.
\end{itemize}

\section{Related Work}
With artificial intelligence moving into the focus of financial service providers, it is now an influential factor in ESG investing \cite{sandp}. The work which has already been done on the topic can be divided into three categories. The first one used artificial intelligence for ESG applications but not for the purpose of exploring the connection between financial and ESG performance. The second part of the research focused on predicting financial performance based on ESG data while the last part had the goal of either predicting ratings or creating entirely new ways of measuring ESG performance. Since many firms publish sustainability reports on an annual basis \cite{teliew2014}, this medium was frequently analyzed by researchers using content analysis or text mining with the aim of identifying topics and trends \cite{freund2013, teliew2014, szeke2017, bjorn2017, moda2009, moda2010, liao2017}. A similar methodology was also applied to shareholder resolutions \cite{raghu2020}.

The information extracted from these reports has been used for various other applications. For instance, text classification algorithms and content analysis were used in order to assess the completeness of sustainability reports \cite{shahi2011}, to relate disclosure of ESG issues to a firm's cost of capital and the information asymmetry between it and its shareholders \cite{michaels2017} and to help managers identify the dimensions where resources should be allocated \cite{lin2018}. Other authors used Twitter data to analyze how corporations should communicate their ESG issues \cite{araujo2018, colleoni2013}.

Apart from the goal, the mentioned papers' main difference to this work consists of the data collected. Most of the researchers primarily focused on textual data which is not taken into consideration here. Correspondingly, the methods used differ as well.

Leaning on the research about the connection between ESG and financial performance \cite{friede:2015}, some authors tried to use machine learning to predict financial performance based on ESG data using regression \cite{suktho2018} or classification approaches \cite{teoh2019}.

Especially, ESG data was used to predict the stock performance of companies \cite{franco2020, hu2018, mitsu2017} or certain financial metrics such as the return of equity (ROE) \cite{lucia2020}. Other authors combined ESG data with the results of an analysis of public sentiment \cite{serafeim2020} or stock returns forecasted by a separate algorithm \cite{vo2019} to reach the same goal.

Although some previous studies used similar data for their analysis as this work and some of them even used comparable algorithms \cite{lucia2020}, the outputs and inputs were reversed. While the aim there was to predict financial performance based on ESG metrics the goal of this study is to predict ESG ratings based on financial and other data. 

In light of the deficiencies of ESG metrics, some researchers tried to create new methods for the analysis of a company's sustainability profile. For example, authors tried to increase interpretability using a fuzzy expert system (FES) \cite{ventu2017} or a fuzzy analytic network process (FANP) \cite{wich2019}. Arguing that the reports analysed by ESG rating agencies were too unreliable, other researchers analysed Twitter data \cite{nema2019} and other online sources using natural language processing (NLP) \cite{Bala2015} or used a heterogeneous information network \cite{hisano:2020} to achieve more accurate ESG ratings. \citeauthor{garcia:2020} \cite{garcia:2020} theorized that certain companies should receive better ESG ratings because of their financial situation. In order to prove this connection, they predicted the ESG ratings of a set of publicly listed companies based on financial measures such as the earnings per share (EPS) or the return on assets (ROA).\par
While these authors tried to find new ways of measuring the handling of ESG issues, there are considerable differences between the approaches. Some partly relied on human experts \cite{ventu2017} which exposes them to the same issues traditional ESG ratings face. Others looked at the problem from an internal perspective \cite{wich2019} or created a binary "bad or not" rating \cite{hisano:2020} not comparable to standard sustainability ratings. Some authors relied exclusively on textual data \cite{Bala2015, nema2019} whose quality is not guaranteed. Finally, \citeauthor{garcia:2020} \cite{garcia:2020} also used a small number of fundamental data features to predict the ESG rating for a subset of the companies analyzed in this work. They used a linear regression model for prediction and restricted their dataset to European companies which might have impacted the generalizability of their results. However, since their experimental setup was the most similar to the one used in this paper, their results will be used as a baseline for our experiments.

\section{Fundamental Data and ESG Ratings} \label{data section}
The data used in this study consists of two parts: the fundamental data which was used as an input to the models and the ESG ratings which were the prediction targets. 
Both parts were sourced from Refinitiv Eikon and are thus are only commercially accessible, however, they still constitute public information in the sense of investment decision-making. 
The data sources coincide regarding neither the timeline nor the number of companies. Therefore, a common intersection was used comprising 7413 companies with annual observations between 2002 and 2019, with some having a shorter time period. 
This section describes both datasets in detail and provides an overview of their properties. 

\subsection{Fundamental Data}
Fundamental data are primarily used by investors in the area of stock analysis in order to decide which stock they should invest in. In this context, fundamental data can be defined as "underlying factors that affect a company’s actual business and its future prospects" \cite{drako2016}. The idea behind this approach is that the stock price of a firm does not show the real value of a company but that over time, the price will adjust until it correctly reflects company's fundamentals \cite{drako2016}.

In the context of this paper, this notion is applied in a similar way. It is theorized that in the long run, the fundamental data of a company should reflect its ESG ratings. Here, the fundamental data primarily consists of financial data and general information.

The general information relates to various non-financial facts. For example, it includes the location of the headquarters, the industry, the index which a company is a part of, the stock exchange where it is listed, or the name of the auditor.

The financial data primarily consists of financial reporting data, i.e., data collected through accounting which are used to prepare financial statements. These provide information about the financial position and performance of a company to numerous different stakeholders, e.g., governments, investors, or employees \cite{melville:2019}.

Normally, large businesses are managed by a small number of directors but owned by many different shareholders who are not involved in the management of their firm. Hence, their only option to obtain information is via its financial statements. Only then are they able to judge whether the businesses they are invested in are adequately managed. Thus, the preparation of financial statements is subject to a number of rules and regulations. These differ across countries and generally stipulate how often financial statements should be published, what content should be included, and which principles they should follow \cite{melville:2019}.\par
Therefore, the financial data are used as input since they are more reliable, more widely available, and more comparable than ESG rating data. In the case of this paper, the financial data include balance sheet items, e.g., assets or equity, information from the income statement, e.g., revenue or profits, and the cash flow statement, e.g., net investing cash flow \cite{refworld:2016}.\par
After pre-processing, the fundamental data consist of 475 numerical and 44 categorical features. Their properties are shown in Table \ref{tab:fundamental_properties}.
\begin{table}[]
    \caption{Properties of the fundamental data}
    \label{tab:fundamental_properties}
    \begin{tabular}{@{}ll@{}}
        \toprule
        \textbf{Property}     &\textbf{Specification}        \\ 
        \midrule
        time period         & 2002-2019 (18 years)      \\
        \# companies        & 7413                   \\
        \# countries        & 84 \\ 
        \# industries       & 10          \\
        \# stock exchanges  & 94    \\    
        \# stock indices    & 56    \\
        \# large cap.       & 1248    \\
        \# mid cap.         & 2741                       \\
        \# small cap.       & 3357         \\
        total market cap.   & USD 63 tril.         \\
        \midrule
    \end{tabular}
\end{table}

\subsection{ESG Data} \label{esgdata}
\begin{figure}[h]
    \centering
    \includegraphics[width=\linewidth]{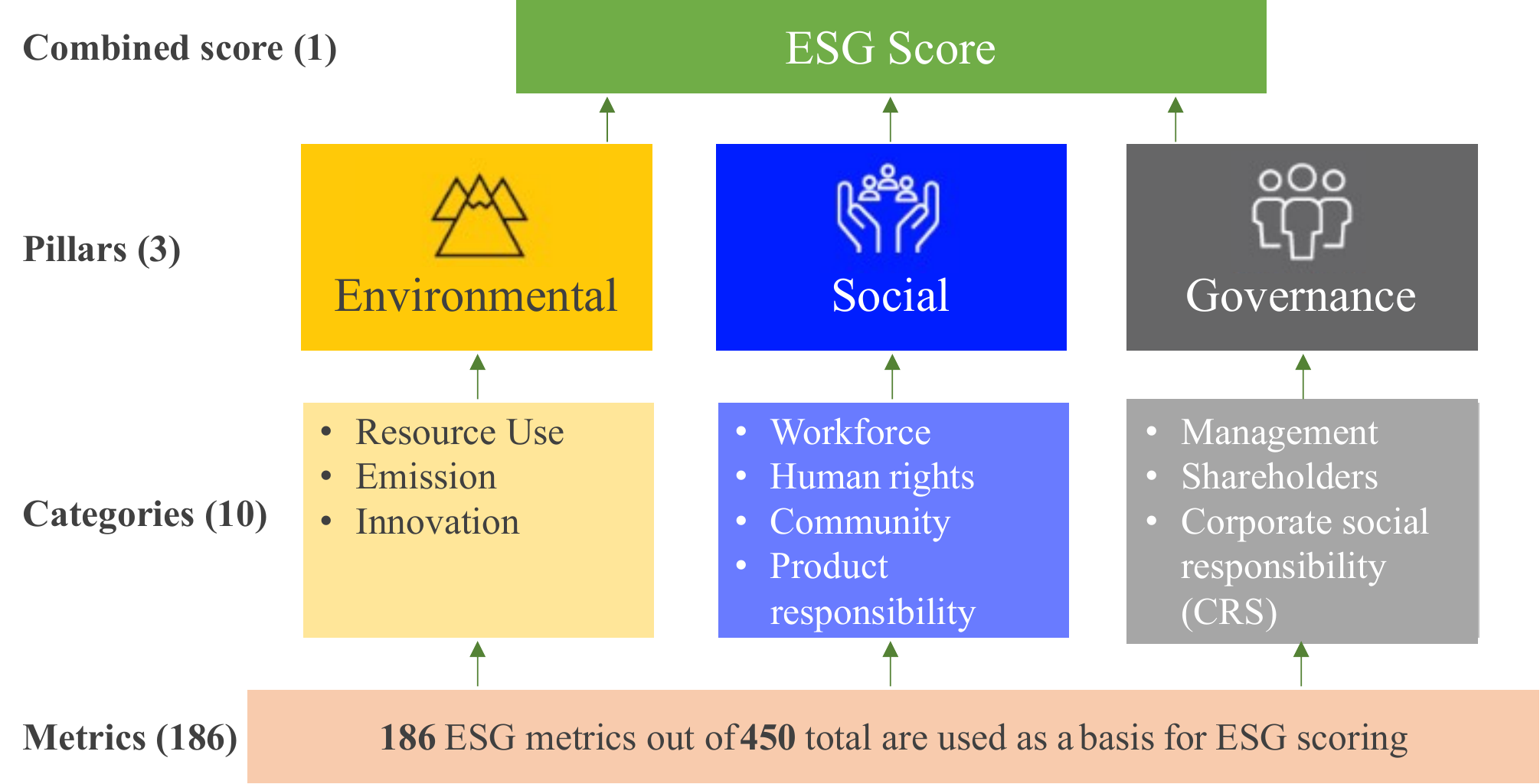}
    \caption{Through analysing sustainability data from various sources 450 metrics are created which represent specific topics such as \(\text{CO}_{2}\) emissions. These are aggregated to form categories, pillars and lastly the overall ESG score.}
    \label{fig:esg_data_hier}
\end{figure}

The rating process by Refinitiv is visualized in Figure \ref{fig:esg_data_hier}.\par

In a first step, it collects data from various sources such as sustainability reports, news sources or NGO websites. It then analyses them to create 450 metrics that measure issues such as the gender pay gap or how much water the company recycled. These are either continuous (e.g., \(\text{CO}_{2}\) emissions) or categorical (e.g., whether a company is involved in the production or sale of firearms). Of those metrics, the 186 most comparable ones (171 categorical, 15 numerical) are used in order to create 10 category ratings listed below.

\begin{description}
    \item[Environmental:] Resource use, emissions, environmental innovation
    \item[Social:] Workforce, human rights, community, product responsibility
    \item[Governance:] Management, shareholders, corporate social responsibility (CSR) strategy
\end{description}

\begin{figure}[h]
    \centering
    \includegraphics[width=\linewidth]{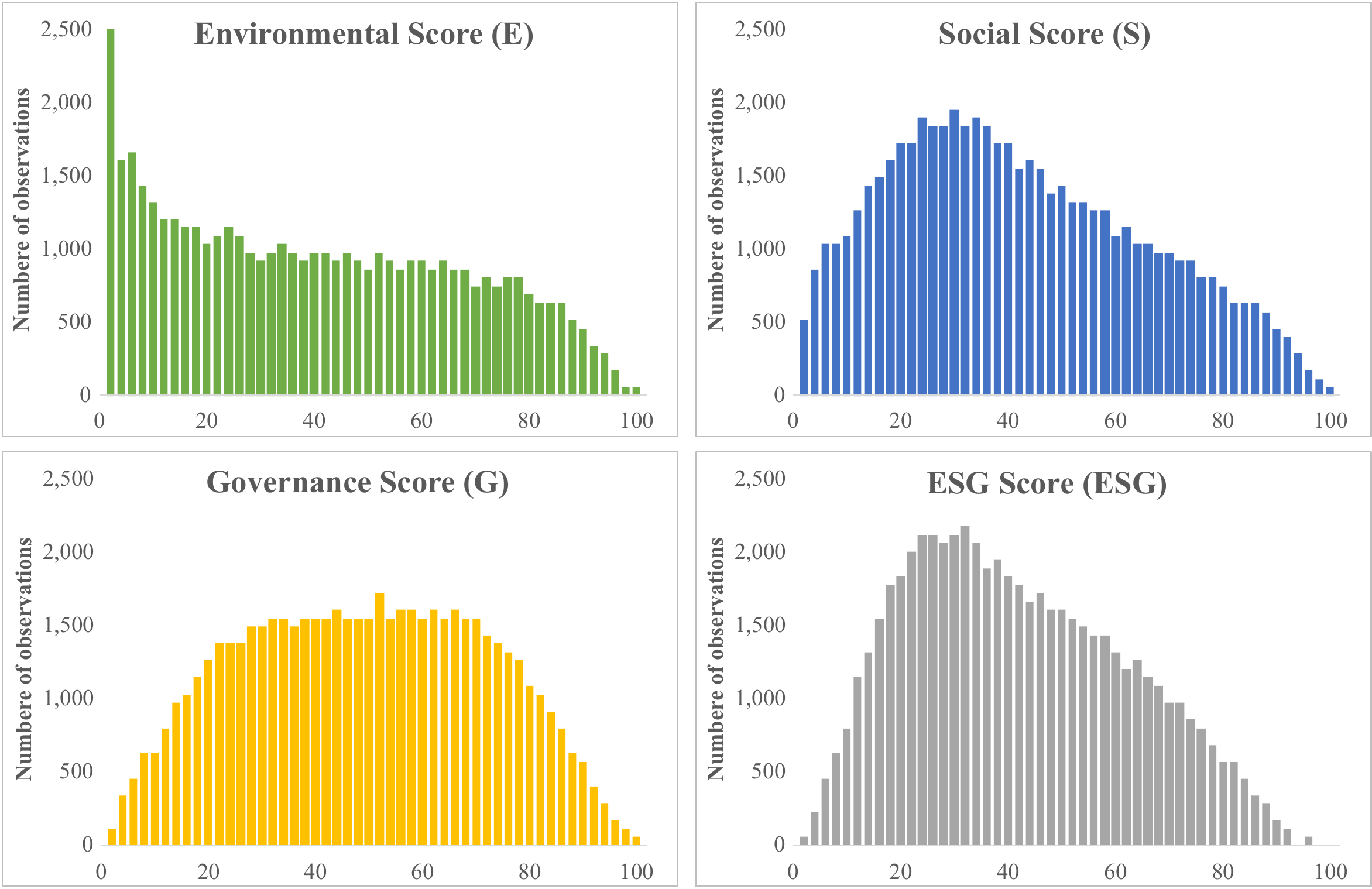}
    \caption{Distribution of the environmental (E), social (S), governance (G) and ESG score. As one can see, the governance (G) rating is symmetric around 50 while the other ratings are positively skewed. The environmental (E) score shows the strongest positive skew with 22.8\% of all values being zero.} 
    \label{fig:hist}
\end{figure}

The categories are grouped in order to form the \textit{Environmental (E)}, \textit{Social (S)}, and \textit{Governance (G)} pillar ratings. Each category rating is a percentile rank score, i.e., they are compared to the performance of the firm's competitors. The category ratings which form the environmental or the social rating are benchmarked against the firms' peers in the same industry while the ones used for the governance rating are benchmarked against the companies which have the same country of incorporation. Finally, the E, S and G ratings are put together to create the \textit{ESG score (ESG)} which combines the information from all different metrics.
Each of the categories, the pillars and the ESG score are rated on a range from 0 (worst) to 100 (best), calculated to two decimal places \cite{refinitiv:2020}. In Figure \ref{fig:hist}, the distributions of the environmental (E), social (S), governance (G) and ESG rating are visualized. The social (S) and ESG ratings show a positively skewed distribution while the governance rating is symmetric around 50. 13097 (22.8\%) of all values of the environmental (E) rating are zero which is not visible due to the y-axis being limited to 2500 companies. This indicates a very strong positive skew.

\section{Approach}
After examining the data set in the previous section, this section will introduce the problem as well as different models and architectures used in this study. 3 different individual machine learning models are used in this work: feedforward neural networks, gradient boosted trees (XGBoost) and categorical gradient boosted trees (CatBoost). Furthermore, 2 model ensembles are employed: a neural network ensemble and a heterogeneous ensemble of 3 individual models.

The goal of this work is to predict different levels of ESG data. As described in section~\ref{esgdata} the combined ESG rating, pillar scores, ESG categories and 15 out of 186 individual metrics are all numerical numbers in the range of 0 to 100. Thus their modeling can be expressed as a regression problem. For the regression task, the mean squared error (MSE) served as the loss function of the models while the $R^2$ (coefficient of determination) and the mean absolute error (MAE) were used to evaluate performance. The remaining 171 categorical metrics were regarded in the classification setting with cross-entropy being the loss function and the accuracy and the F1 score used as evaluation metrics.

\textbf{Feedforward Neural Network}. 
The first individual model used is a feedforward neural network \cite{rumelhart1986learning} with fully connected layers, since the data does not exhibit any internal structure and thus would not benefit from kernels or convolutional layers.
Different architectures are employed using dropout, l1 and l2 regularization as well as batch normalization. For categorical input, embedding layers are optionally added aimed at improving the performance of the model \cite{SAE:2018} as well as its training speed. 

The embedding layers are learned end-to-end with the network and are tasked with reducing the number of dimensions of one-hot encoded categorical variables, thus reducing the number of parameters and computational cost of the model, as well as having regularizing effect. Each categorical feature $f_i$ has its own embedding layer $emb_i$ which only receives $f_i$ as an input. After being passed through the activation function the outputs of all $emb_i$ are concatenated with each other as well as with unchanged numerical inputs of the model and are then passed to hidden layers of the network. This architecture achieves superior results in our testing as will be discussed in the next section.

\textbf{CatBoost} \cite{prokho:2018} ---
the next individual model is a type of gradient boosting trees \cite{drucker_boosting_1995} that need minimal pre-processing since it can natively handle categorical features. For those which only include few categories (controlled by hyper-parameter), CatBoost uses one-hot encoding, while for those with many, ordered target encoding is employed, where each category $x_k^i$ of the $k$-th training vector is replaced with the target statistic (TS) $\hat{x}_k^i \approx \mathbb{E}(y|x^i = x_k^i)$. Ordered boosting with one-side gradient-based is used for optimization.

\textbf{XGBoost} \cite{xgboost} ---
a second tree boosting model used adopting a greedy algorithm that uses the second-order Taylor series and improves generalization via regularization terms. Unlike CatBoost categorical are not handles as a part of the algorithm and have to be one-hot-encoded. However, depending on the dataset, XGBoost can often outperform CatBoost and does also provide sufficiently different predictions that may improve the diversity of the ensemble model presented below.

\textbf{Naive NN Ensemble}.
One of the advantages of CatBoost and XGBoost over neural networks is that they are implicitly combining weak learners into strong ones. Similarly, to improve NN model performance conventional naive ensemble of 10 networks was used in this study. It has been shown empirically and theoretically that ensembling techniques improve the generalization performance of neural networks \cite{levin_statistical_1990}, due to them ending in different local minima during training and thus capturing different data modes \cite{hansen_neural_1990}. The aggregation method for ensemble members was chosen to be median, due to its robustness to outlier predictions \cite{lakshmi:2017}. 

\textbf{Heterogeneous Ensemble}. It is well known that member diversity improves ensemble performance \cite{kuncheva_measures_2003}. One way to improve diversity is to use structurally different models \cite{tahir_multilabel_2012}. The main reasoning behind it is that members can learn different aspects of the dependence between input and output, and can often in combination outperform every single model in it \cite{ren:2016}. Therefore, the last heterogeneous ensemble model is computed based on the median of predictions of Catboost, XGBoost, and a single neural network to leverage their distinct architectural characteristics.

\section{Empirical Evaluation}
In this section, we present our in-depth empirical evaluation of our individual models and model ensembles. We describe the experimental setup, define the baselines, and present our results.

\subsection{Experimental setup}
In Figure~\ref{fig:setup}, the general setup of this work is presented.
All the experiments were performed for a single target chosen from ESG scores in a regression setting for continuous, or classification setting for categorical targets. 

Since the aim of this study is to predict ESG scores with only fundamental data, previous ESG scores could not be used, thus each year of a company was regarded as a different entity, which resulted in a total number of 57310 observations. Temporal dependencies in fundamental data were not modelled to enable creation of ratings for newly listed companies.
Due to the high correlation between scores for the same company across years, the train-validation-test split (60\%-20\%-20\%) was done in a way to include each company in only one part of the split.

As part of the pre-processing, the dataset was cleaned for collinear features as well as high cardinality ones to avoid possible memorization. In addition, the categories of every categorical feature which only included one company were aggregated as "other" in order to reduce the model complexity and prevent the possibility to identify firms individually.
Input data were also imputed, scaled for numerical, and encoded for categorical features.

Architecture and hyper-parameters grid search were then employed on the validation set for the ESG target in order to find the best performing models specification. The combinations of parameters tested for neural networks are presented in Table~\ref{tab:nn_gridsearch}, 

\begin{table}[]
    \caption{Grid search for neural network architectures}
    \label{tab:nn_gridsearch}
    \begin{tabular}{@{}ll@{}}
        \toprule
        \textbf{Model parameter}\phantom{aaaaa}  &\textbf{Value combinations}\phantom{aaaaaa}  \\ 
        \midrule
        \# layers (depth)   & 4, 5, 6                   \\
        \# neurons (width)  & 2000, 1200, 500, 250      \\
        activation function & Sigmoid, ReLu, Leaky ReLu \\ 
        dropout             & 0, 0.3, 0.5, 0.7          \\
        learning rate       & $10^{-3}$, $10^{-4}$, $10^{-5}$    \\    
        weight decay        & 0.01, 0.03, 0.1, 0.4, 1, 2, 10    \\
        l1 rate             & 0.05, 0.1, 0.2, 0.4, 1, 10    \\
        batch normalization & no, yes                       \\
        embeddings          & no, per country, per industry         \\
        embeddings size     & 1, 2, 3, 4, 10\% of categories         \\
        epochs              & 1000, 2000, 3000, 5000, 15000   \\      
        early stopping      & no, yes   \\   
        \midrule
    \end{tabular}
\end{table}

\begin{table}[]
    \caption{Grid search for CatBoost architectures}
    \label{tab:cb_gridsearch}
    \begin{tabular}{@{}ll@{}}
        \toprule
        \textbf{Model parameter}\phantom{aaaaa}  &\textbf{Value combinations}\phantom{aaaaaa}  \\ 
        \midrule
        Tree depth          & 3, 4, 5, 10               \\
        learning rate       & $10^{-1}$, $10^{-2}$, $10^{-3}$    \\    
        max \# one-hot-encoding        & 10, 30, 255    \\
        iterations          & 1000, 2000, 5000, 10000   \\ 
        early stopping      & no, yes   \\   
        \midrule
    \end{tabular}
\end{table}

\begin{table}[]
    \caption{Grid search for XGBoost architectures}
    \label{tab:xgb_gridsearch}
    \begin{tabular}{@{}ll@{}}
        \toprule
        \textbf{Model parameter}\phantom{aaaaa}  &\textbf{Value combinations}\phantom{aaaaaa}  \\ 
        \midrule
        Max. tree depth          & 3, 5, 10, 15              \\
        learning rate       & $10^{-1}$, $10^{-2}$, $10^{-3}$    \\    
        min. child weight        & 0.5, 0.7, 1    \\
        number of trees          & 100, 200, 300   \\ 
        early stopping      & no, yes   \\   
        \midrule
    \end{tabular}
\end{table}

\begin{figure}[h]
    \centering
    \includegraphics[width = \linewidth]{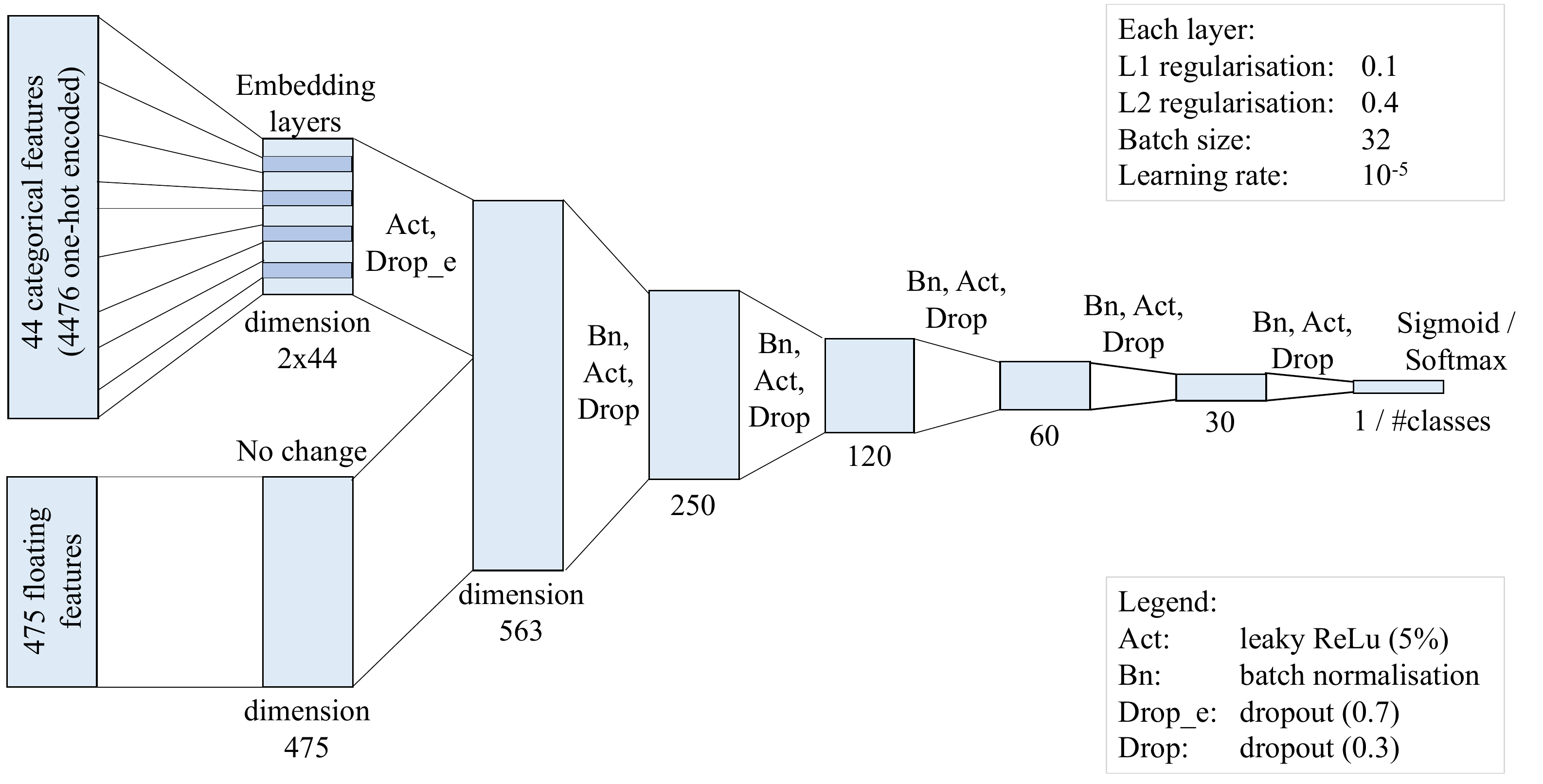}
    \caption{Tuned feedforward neural network architecture. Categorical features first pass through separate encoding layers before being recombined with numerical ones.}
    \label{fig:nn_architecture}
\end{figure}

After conducting hyper-parameter searches the following specifications were chosen: 

\begin{itemize}
    \item \textbf{Feedforward neural network} (6 layers, with categorical embedding, exact architecture is depicted in the Figure~\ref{fig:nn_architecture})
    \item \textbf{CatBoost} (iterations: 10000, learning rate: 0.01, tree depth: 4, one-hot-encoding max. size: 255, early stopping: yes)
    \item \textbf{XGBoost} (number of trees: 200, learning rate: 0.1, tree depth: 10, min. child weight: 1, early stopping: yes)
\end{itemize}

As a part of each experiment, individual model architectures (feedforward neural network, CatBoost, and XGBoost) were fitted on the train set with early stopping based on the validation set. Model predictions were then combined in a heterogeneous ensemble as the median for continuous and majority voting for categorical targets. Additionally, 10 more neural networks of the same architecture were independently trained and combined in a naive NN Ensemble using the same aggregation method. 
The experiments have been conducted on NVIDIA Tesla V-100 16GB GPUs paired with 3.3GHz Intel CPUs. The overall time for 1 experiment for a single target took approximately 1.5 days.

\subsection{Baselines results}
For the evaluation of our models, we compared our results to two baselines.
\begin{itemize}
    \item \textbf{Baseline 1:} Mean rating per industry. 
    \item \textbf{Baseline 2:} Linear model for ESG score by \citeauthor{garcia:2020} \cite{garcia:2020}.
\end{itemize}
For our first baseline, we calculated the mean ratings per industry since we considered this to be more accurate than calculating the mean over all companies. For example, the predictions for the ESG scores achieved this way showed an $R^2$ of 6\% and an MAE of 16.5. The baseline 1 results for all targets are shown as red lines in Figures~\ref{fig:results1} and \ref{fig:results2}, and are listed in Table \ref{tab:mae_results}.
\citeauthor{garcia:2020} \cite{garcia:2020} used a different universe of data (only publicly listed, European companies covering 2013-2018, only 8 features, etc.) and reported only $R^2$ for the combined ESG score of 27\% on the test set. This baseline is depicted in Figure~\ref{fig:results1} as a blue line. Since the exact data selection criteria were not published we could not replicate the results. Using our larger dataset and following the approach of García et al. we were able to achieve only 16\% $R^2$ for the ESG score while we arrived at $R^2$ of 33\% using a CatBoost model. This indicates that ESG data exhibit certain non-linear relationships with fundamental data and the use of advanced machine learning models is justified.

\subsection{Approach results}
\begin{figure}[h]
    \centering
    \includegraphics[width=\linewidth]{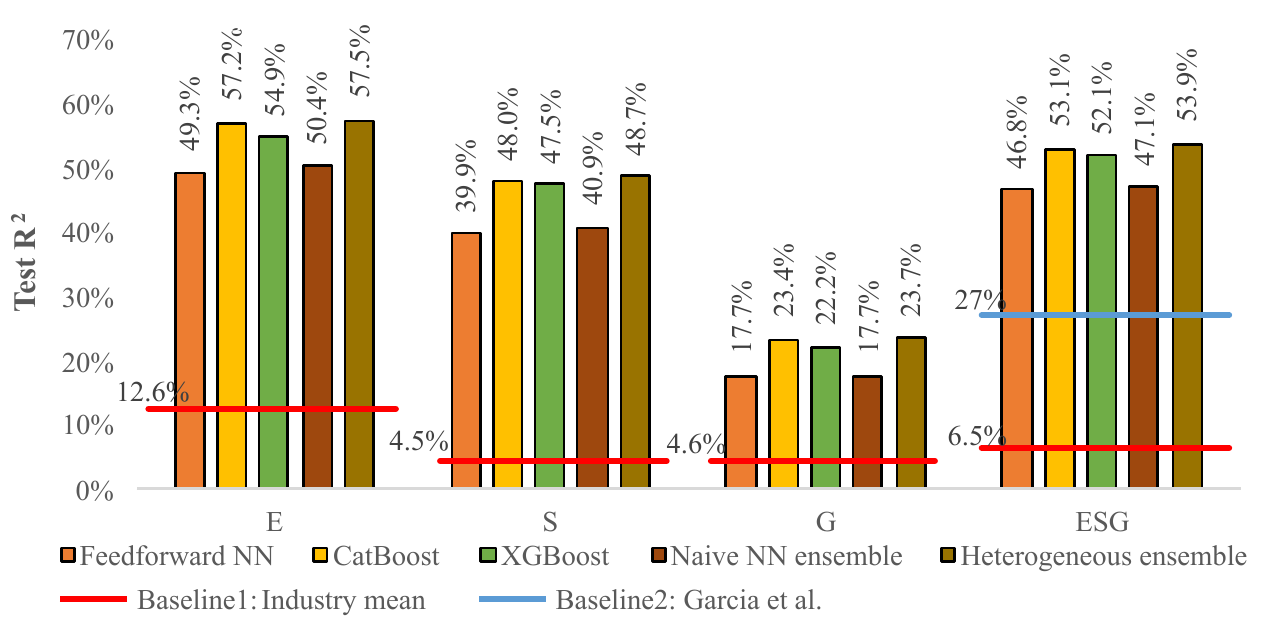}
    \caption{Test $R^2$ for the pillar ratings (E, S, G) and the overall ESG score prediction. Governance (G) rating seems to be much harder to predict.}
    \label{fig:results1}
\end{figure}

In this section, the results of the empirical study are presented. Figure~\ref{fig:results1} shows the accuracies of the prediction models as measured by $R^2$ score for the 3 pillar scores (E, S, G) and combined ESG scores, while Table~\ref{tab:mae_results} lists the corresponding MAE. We can see that CatBoost has performed the best compared to other individual models, followed by XGBoost and NNs. We attribute the success of CatBoost to its special way to handle categorical variables. Looking at the ensemble performance we can identify that the Naive NN ensemble has indeed improved the performance of the individual NN model, however, this was not enough to catch up with gradient boosted tree models. The heterogeneous ensemble, on the other hand, has the best performance on the test set for all targets and has outperformed all of its constituencies. We attribute its success by it combining the strength of each model family in finding distinct relationships between input and target data. 

Overall, one can see that predicting the \textit{environmental} rating was the easiest, followed by the \textit{ESG and social} rating. Interestingly, \textit{governance} rating was the hardest to model. We attribute this to relatively less information on governance structure contained in structured financial and fundamental data. For the ESG score heterogeneous ensemble was able to explain almost 54\% of the variance and bring mean absolute error to 11.3 percentage points, which can already be used as a good basis for the creation of initial company scores in an automated fashion using easily available data. Table~\ref{tab:sp500_2} illustrates the ESG and pillar score prediction accuracy based on randomly selected companies belonging to S\&P 500 index (500 biggest companies in the USA). 

\begin{table}[]
    \caption{Mean Absolute Error (MAE) for ESG, pillars and categories prediction using baseline1 (B1), feedforward NN (FNN), CatBoost (CB), XGBoost (XGB), NN ensemble (NNE) and heterogeneus ensemble (HE).}
    \label{tab:mae_results}
    \begin{tabular}{@{}lllllll@{}}
    \toprule
                  & \textbf{B1} & \textbf{FNN} & \textbf{CB} & \textbf{XGB} & \textbf{NNE} & \textbf{HE} \\ \midrule
    ESG Score       & 16.5  & 12.1  & 11.3     & 11.4    & 12.1    & \textbf{11.2}        \\
    E Score         & 22.7  & 16.3  & 15.0     & 15.2    & 16.2    & \textbf{14.9}        \\
    S Score         & 19.0  & 14.6  & \textbf{13.5}     & 13.6    & 14.5    & \textbf{13.5}   \\
    G Score         & 18.7  & 17.4  & \textbf{16.7}     & 16.8    & 17.4    & \textbf{16.7}  \\
    \midrule
    Resource Use    & 27.9  & 20.0  & 18.5     & 18.6    &         & \textbf{18.4}        \\
    Emissions       & 27.8  & 19.6  & \textbf{18.0}     & 18.4    &         & \textbf{18.0}        \\
    Env. Innovation & 21.4  & 18.2  & 17.8     & 17.6    &         & \textbf{17.5}        \\
    Workforce       & 24.6  & 19.5  & \textbf{17.3}     & 17.9    &         & 17.6        \\
    Human Rights    & 24.5  & 18.0  & 17.2     & 17.1    &         & \textbf{16.9}        \\
    Community       & 24.7  & 20.0  & \textbf{18.9}     & 19.1    &         & \textbf{18.9}        \\
    Product Resp.   & 26.3  & 23.6  & \textbf{22.6}     & 22.8    &         & \textbf{22.6}        \\
    Management      & 24.3  & 23.1  & \textbf{22.4}     & 22.6    &         & 22.5        \\
    Shareholders    & 24.5  & 24.3  &\textbf{23.8}     & 23.9    &         & 23.9        \\
    CSR Strategy    & 26.7  & 20.3  & 19.2     & 19.2    &         & \textbf{18.9}        \\ \bottomrule
    \end{tabular}
\end{table}

\begin{figure}[h]
    \centering
    \includegraphics[width=\linewidth]{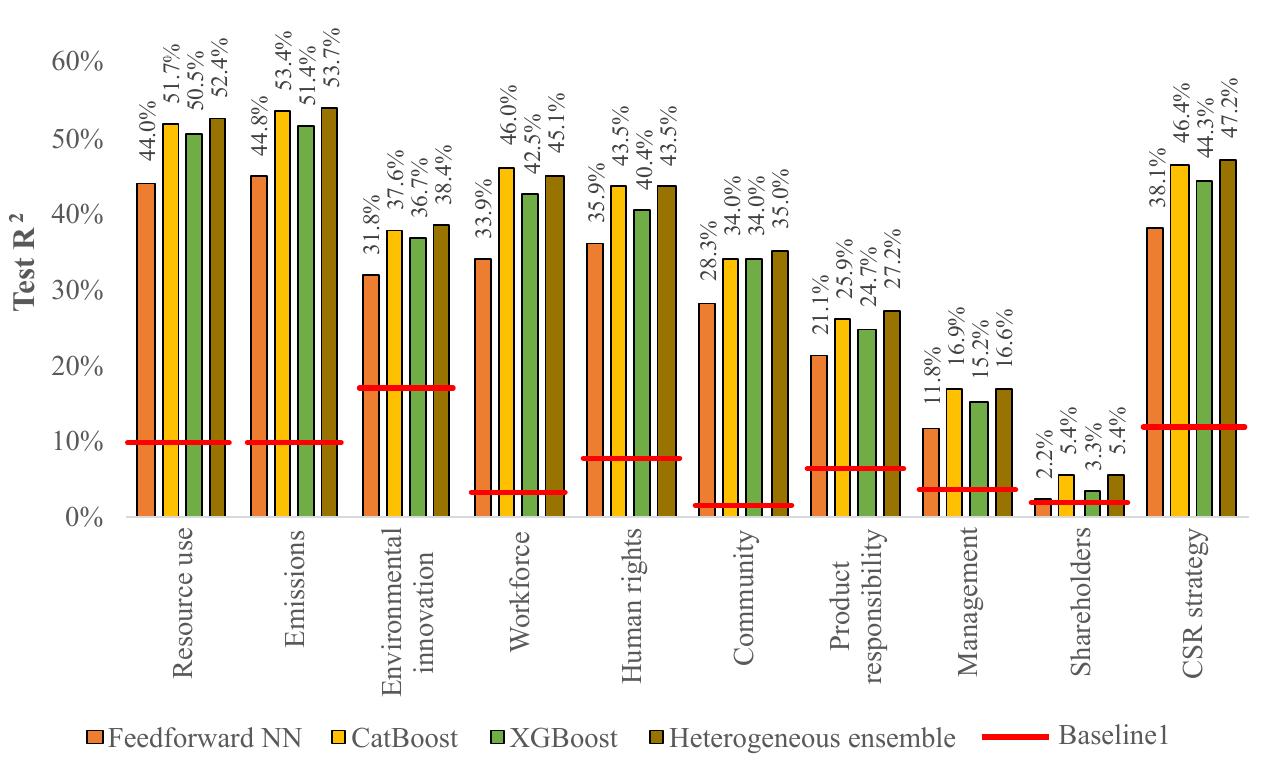}
    \caption{Test $R^2$ for the 10 ESG categories prediction. Governance related categories are again much harder to predict.}
    \label{fig:results2}
\end{figure}

After examining the highest levels of ESG data, it was then analyzed how well one could predict the 10 ESG categories. Since the creation of feedforward NN ensembles was computationally expensive and did not improve final results for pillars and overall rating, it was not done here. Figure~\ref{fig:results2} presents the $R^2$ on the test set for the 10 ESG categories and MAE numbers can be found in Table~\ref{tab:mae_results}. The results are similar to the top-level prediction with categories belonging to the \textit{environmental} pillar (\textit{resource Use}, \textit{emissions} and \textit{environmental innovation}) as well as the \textit{social} pillar (\textit{workforce}, \textit{human rights}, \textit{community}, and \textit{product responsibility}) showing relatively higher performance, while those belonging to the \textit{governance} pillar (\textit{management}, \textit{shareholders} and \textit{corporate social responsibility strategy}) proving much harder to predict. Similarly to the top-level heterogeneous ensemble outperformed all the individual models (except for single category \textit{workforce}) highlighting the power of aggregating structurally different models.

\begin{figure}[h]
    \centering
    \includegraphics[width=\linewidth]{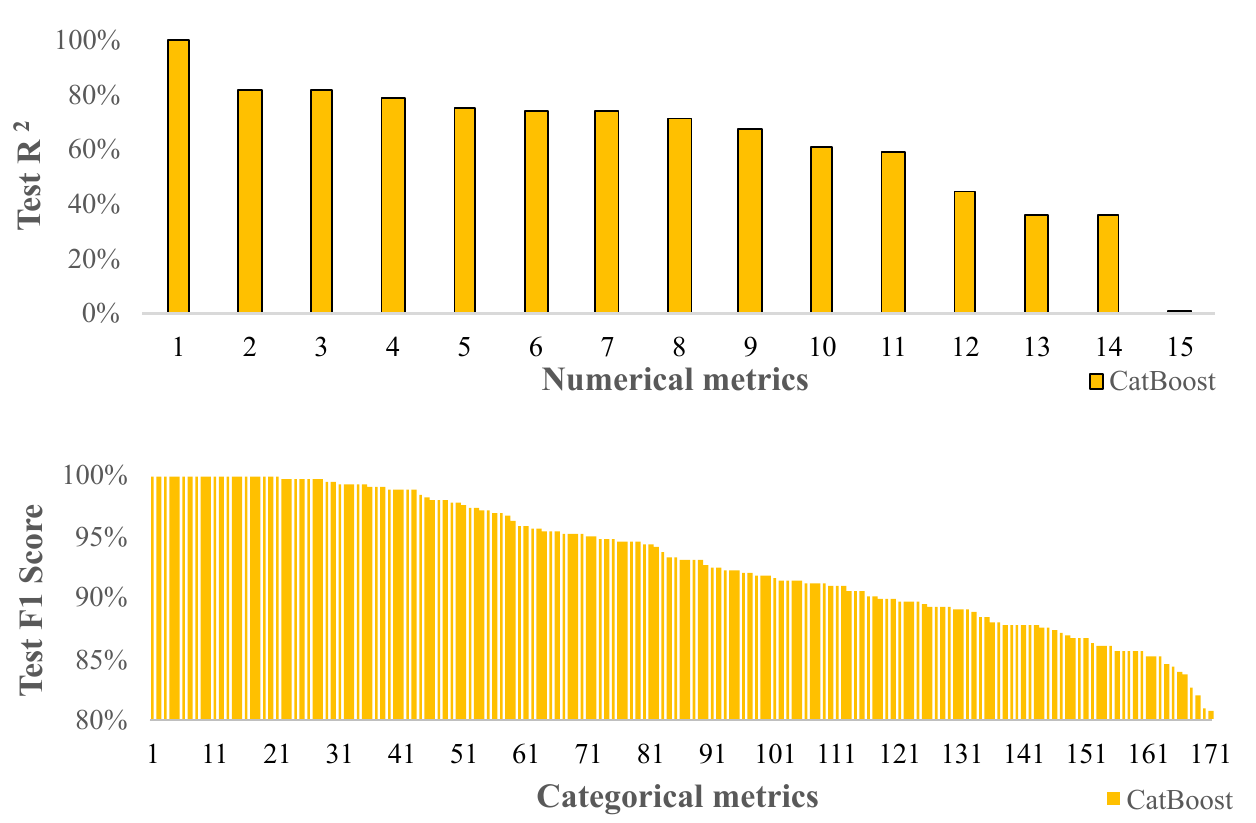}
    \caption{Sorted test $R^2$ and F1-scores of CatBoost model for individual ESG metrics. Numerical metrics exhibit similar performance to higher level targets, while categorical all have F1 score >80\%.}
    \label{fig:results3}
\end{figure}

The experiments were also performed for some of the individual ESG metrics as targets and generally showed a similar pattern with ensembles generally outperforming individual models. However, due to computational cost, only the single best individual model (CatBoost) was used for all individual metrics. Due to the sheer number of targets, the results will not be presented here in detail. However, for the 15 numerical metrics, the average $R^2$ was in the range of 62\% and for categorical metrics, the average F1-score was around 93\%. The visualization of these results can be seen in the Figure~\ref{fig:results3}

\section{Conclusion}
Investors increasingly consider sustainability aspects as important for investment decisions. They have started to assess how companies handle such topics. To support this process, specialized rating agencies supply a number of metrics relating to environmental, social and governance (ESG) dimensions.

However, ESG ratings are based on voluntarily disclosed sustainability reports and subjective judgments. Ratings created by different providers can differ substantially and due to proprietary methodology reconciliation is usually impossible. Since the methodology is labor-intensive, the creation of ratings is a time-consuming process. Furthermore, smaller companies often do not publish sustainability reports. This limits the universe of assessed companies for investment decisions.

This work proposed to overcome the aforementioned issues by creating automated ESG ratings based on publicly available data without relying on sustainability reporting. This was done by training heterogeneous ensemble models using the strengths of neural networks and gradient boosted trees. The input consisted of fundamental data, i.e., data pertaining to the industry, geographical location, and financial performance. Our most accurate ensemble model was constructed using Catboost, XGBoost and a feedforward neural network and allowed us to explain 54\% of the variance in the overall ESG score ($R^2$), with a mean absolute error of 11.2 percentage points. The level of explainability for pillar scores (E, S and G) and individual metrics such as \(\text{CO}_{2}\) emissions or gender diversity was high as well. Our results outperformed the baselines and prior work we have identified for the problem.

With the help of our model, initial ESG ratings can be created for every publicly traded company in an automated way. 
Such ratings can either be used by investors directly or by rating providers as a basis to speed up the rating process as well as issue ratings for companies without sustainability reporting, allowing them to improve coverage.

We believe our work can significantly increase ESG data availability, and help investors to identify companies with superior environmental, social and governance practices. 
This can attract capital and support such companies advancing the sustainability of the society overall.


\begin{acks}
The data used for this paper were sourced from the Eikon platform by Refinitiv. In accordance with its usage rights policy, only derived and aggregated results are shown.
\end{acks}

\newpage

\onecolumn

\begin{table}[h]
\caption{Environmental (E), social (S), governance (G) and ESG ratings predictions using heterogeneous ensemble model for 50 randomly selected companies from S\&P 500 index in the test set. Company names anonymized.}
\label{tab:sp500_2}
\resizebox{\textwidth}{!}{%
\begin{tabular}{@{}lllllllll@{}}
\toprule
                                      & \multicolumn{2}{c}{\textbf{ESG}}    & \multicolumn{2}{c}{\textbf{Environmental (E)}} & \multicolumn{2}{c}{\textbf{Social (S)}} & \multicolumn{2}{c}{\textbf{Governance (G)}} \\ \midrule
                                      & predicted & abs. error & predicted     & abs. error    & predicted & abs. error & predicted   & abs. error   \\
Consumer Cyclicals Company A         & 59.0      & 6.7            & 59.2          & 2.0               & 61.7      & 27.4           & 46.9        & 27.5             \\
Financials Company A                 & 52.2      & 4.6            & 36.1          & 18.9              & 62.3      & 15.1           & 57.6        & 18.2             \\
Consumer Cyclicals Company B         & 49.6      & 4.8            & 29.5          & 6.3               & 52.4      & 23.3           & 51.6        & 29.3             \\
Technology Company A                 & 65.5      & 5.2            & 64.1          & 0.4               & 72.2      & 16.4           & 57.5        & 0.2              \\
Energy Company A                     & 61.1      & 21.4           & 61.9          & 32.7              & 71.1      & 19.9           & 44.6        & 11.5             \\
Energy Company B                    & 68.2      & 14.7           & 69.5          & 17.6              & 78.2      & 15.2           & 64.1        & 1.1              \\
Energy Company C                    & 60.5      & 0.7            & 55.9          & 4.8               & 64.3      & 5.0            & 68.5        & 3.1              \\
Financials Company B                 & 48.6      & 29.7           & 30.8          & 49.7              & 58.1      & 30.3           & 58.4        & 8.1              \\
Technology Company B                 & 55.5      & 0.2            & 45.6          & 0.9               & 56.3      & 2.5            & 58.8        & 3.7              \\
Consumer Cyclicals Company C         & 32.8      & 2.2            & 10.7          & 10.7              & 35.8      & 0.9            & 44.1        & 6.7              \\
Healthcare Company A                 & 47.1      & 18.8           & 26.3          & 13.7              & 43.2      & 7.1            & 50.7        & 25.3             \\
Financials Company C                 & 48.3      & 10.2           & 35.1          & 3.9               & 54.3      & 11.4           & 47.5        & 10.8             \\
Financials Company D                 & 49.2      & 5.1            & 25.6          & 17.2              & 53.5      & 1.1            & 48.4        & 19.4             \\
Utilities Company A                  & 61.2      & 2.5            & 60.5          & 5.5               & 57.9      & 5.6            & 67.8        & 7.8              \\
Industrial Company A                 & 29.4      & 8.4            & 22.6          & 21.4              & 36.3      & 9.9            & 32.7        & 5.3              \\
Industrial Company B                 & 53.2      & 17.9           & 50.5          & 28.1              & 60.4      & 32.6           & 48.5        & 15.2             \\
Technology Company C                 & 57.1      & 17.3           & 50.6          & 42.8              & 59.4      & 6.9            & 55.0        & 14.1             \\
Technology Company D                 & 46.8      & 0.4            & 19.5          & 7.5               & 49.7      & 1.9            & 44.1        & 13.1             \\
Consumer Cyclicals Company D         & 52.9      & 7.5            & 45.5          & 45.1              & 55.6      & 10.2           & 60.9        & 5.2              \\
Consumer Cyclicals Company E         & 62.5      & 12.9           & 58.5          & 1.6               & 70.2      & 13.7           & 56.7        & 18.4             \\
Consumer Cyclicals Company F         & 26.7      & 26.2           & 3.9           & 68.2              & 33.7      & 11.4           & 34.4        & 9.8              \\
Industrial Company C                  & 74.5      & 10.1           & 73.0          & 10.2              & 81.5      & 15.2           & 65.9        & 1.1              \\
Financials Company E                 & 37.4      & 11.0           & 14.4          & 11.9              & 38.1      & 5.9            & 49.2        & 23.5             \\
Consumer Cyclicals Company G         & 74.6      & 8.0            & 74.2          & 7.2               & 74.1      & 17.2           & 69.7        & 0.4              \\
Utilities Company B                  & 63.2      & 2.4            & 62.5          & 14.2              & 65.8      & 9.0            & 63.8        & 5.6              \\
Financials Company F                 & 53.5      & 10.6           & 36.6          & 40.6              & 59.0      & 4.7            & 58.6        & 2.0              \\
Industrial Company D                 & 67.9      & 1.3            & 64.9          & 15.0              & 68.9      & 1.1            & 63.3        & 19.1             \\
Consumer Non-Cyclicals Company H     & 52.8      & 15.9           & 48.3          & 9.2               & 56.8      & 38.0           & 41.8        & 5.6              \\
Telecommunications Services Company A & 70.5      & 5.4            & 75.3          & 22.4              & 70.1      & 2.1            & 69.6        & 9.4              \\
Financials Company G                 & 44.1      & 25.7           & 14.5          & 27.7              & 49.7      & 25.5           & 50.3        & 23.2             \\
Financials Company H                & 65.7      & 11.7           & 60.4          & 1.3               & 69.3      & 8.4            & 61.5        & 22.9             \\
Financials Company I                  & 67.2      & 6.4            & 58.9          & 32.9              & 73.3      & 10.5           & 69.8        & 10.2             \\
Technology Company E                 & 56.2      & 19.3           & 36.5          & 38.9              & 56.3      & 19.6           & 52.7        & 22.4             \\
Financials Company J                & 66.4      & 11.5           & 66.2          & 22.5              & 66.5      & 20.1           & 69.1        & 1.2              \\
Consumer Cyclicals Company I         & 52.9      & 29.5           & 37.2          & 45.5              & 69.5      & 16.0           & 63.1        & 14.3             \\
Technology Company F                 & 47.1      & 9.9            & 44.2          & 31.0              & 52.2      & 12.6           & 42.9        & 2.0              \\
Technology Company G                 & 55.3      & 3.8            & 40.0          & 5.6               & 55.5      & 3.8            & 50.5        & 5.9              \\
Utilities Company C                  & 68.6      & 0.5            & 62.1          & 16.6              & 68.9      & 13.6           & 69.8        & 1.0              \\
Financials Company K                 & 42.9      & 0.5            & 13.8          & 13.8              & 40.0      & 9.3            & 47.7        & 23.0             \\
Healthcare Company B                 & 65.5      & 12.3           & 62.7          & 8.5               & 71.0      & 20.1           & 59.6        & 3.2              \\
Healthcare Company C                 & 49.7      & 25.2           & 35.3          & 26.7              & 53.8      & 29.1           & 54.6        & 21.2             \\
Technology Company H                 & 40.4      & 23.4           & 15.8          & 46.6              & 55.1      & 25.6           & 44.3        & 5.4              \\
Technology Company I                  & 59.9      & 16.1           & 55.5          & 35.8              & 66.8      & 2.4            & 59.2        & 17.7             \\
Basic Materials Company A            & 55.5      & 5.2            & 53.9          & 2.4               & 54.2      & 0.6            & 54.6        & 26.2             \\
Energy Company D                     & 80.9      & 12.0           & 83.0          & 6.8               & 82.1      & 18.1           & 75.8        & 29.0             \\
Utilities Company D                  & 61.4      & 8.0            & 54.3          & 3.9               & 57.3      & 22.6           & 68.5        & 1.2              \\
Consumer Non-Cyclicals Company J     & 64.2      & 1.8            & 61.0          & 15.2              & 70.7      & 4.4            & 58.0        & 17.1             \\
Financials Company L                 & 42.6      & 5.7            & 21.4          & 1.1               & 53.5      & 30.2           & 55.8        & 13.2             \\
Basic Materials Company B            & 67.1      & 9.6            & 70.8          & 9.4               & 66.4      & 3.3            & 63.5        & 19.1             \\
Healthcare Company D                & 43.1      & 14.1           & 23.5          & 40.6              & 50.8      & 45.8           & 39.9        & 18.9             \\ \bottomrule
\end{tabular}}
\end{table}

\twocolumn

\bibliographystyle{ACM-Reference-Format}
\bibliography{main.bib}



\end{document}